%% file: gaptv_icml.tex
\documentclass{article}

\usepackage{times}
\usepackage{graphicx} % more modern
\usepackage{natbib}

\usepackage{algorithm}
\usepackage{algorithmic}

\usepackage{hyperref}

\usepackage{amsmath,amssymb,amsthm,amsfonts}
\usepackage{caption,subcaption}
\usepackage{xr}
\usepackage{url}
\usepackage{amssymb}
\usepackage{amsbsy}
\usepackage{amsthm}
\usepackage{amsmath}
\usepackage{paralist}
\usepackage{bm}
\usepackage{booktabs}
\usepackage{rotating}
\usepackage{cases}
\usepackage{mathtools}
\usepackage{setspace}
\usepackage{textcomp}
\usepackage{nicefrac}       % compact symbols for 1/2, etc.
\usepackage{microtype}      % microtypography

\usepackage[accepted]{whi2017}

\newcommand{\R}{\mathbb{R}}
\newcommand{\vnorm}[1]{\left|\left|#1\right|\right|}

\icmltitlerunning{Interpretable Low-Dimensional Regression}

\begin{document} 

\twocolumn[
\icmltitle{Interpretable Low-Dimensional Regression via Data-Adaptive Smoothing}



% you can specify symbols, otherwise they are numbered in order
\icmlsetsymbol{equal}{*}

\begin{icmlauthorlist}
\icmlauthor{Wesley Tansey}{uta,cumc}
\icmlauthor{Jesse Thomason}{uta}
\icmlauthor{James G. Scott}{uta}
\end{icmlauthorlist}

\icmlaffiliation{uta}{University of Texas at Austin, Austin, Texas, USA}
\icmlaffiliation{cumc}{Columbia University Medical Center, New York, New York, USA}

\icmlcorrespondingauthor{Wesley Tansey}{tansey@cs.utexas.edu}

\icmlkeywords{interpretability, smoothing, convex regression}

\vskip 0.3in
]



\printAffiliationsAndNotice{}  % leave blank if no need to mention equal contribution
 
\begin{abstract} 
\input{abstract}
\end{abstract} 

\input{introduction}

\input{background}

\input{algorithm}

\input{evaluation}

\bibliography{gaptv_icml}
\bibliographystyle{icml2017}

\end{document}

%% file: abstract.tex
% !TEX root = gaptv_icml.tex
We consider the problem of estimating a regression function in the common situation where the number of features is small, where interpretability of the model is a high priority, and where simple linear or additive models fail to provide adequate performance. To address this problem, we present Maximum Variance Total Variation denoising (MVTV), an approach that is conceptually related both to CART and to the more recent CRISP algorithm \citep{petersen:etal:2016}, a state-of-the-art alternative method for interpretable nonlinear regression. MVTV divides the feature space into blocks of constant value and fits the value of all blocks jointly via a convex optimization routine. Our method is fully data-adaptive, in that it incorporates highly robust routines for tuning all hyperparameters automatically. We compare our approach against CART and CRISP via both a complexity-accuracy tradeoff metric and a human study, demonstrating that that MVTV is a more powerful and interpretable method.%\footnote{A full version of this paper is currently in submission to \textit{NIPS'17}.}

%% file: introduction.tex
% !TEX root = gaptv_icml.tex
\section{Introduction}
\label{sec:introduction} 
A recent line of research in interpretable machine learning focuses on low-dimensional regression, where the feature set is relatively small and human intelligibility as a primary concern. For example, lattice regression with monotonicity constraints has been shown to perform well in video-ranking tasks where interpretability was a prerequisite \cite{gupta:etal:2016}. The interpretability of the system enables users to investigate the model, gain confidence in its recommendations, and guide future recommendations. In the two- and three- dimensional regression scenario, the Convex Regression via Interpretable Sharp Partitions (CRISP) method \cite{petersen:etal:2016} has recently been introduced as a way to achieve a good trade off between accuracy and interpretability by inferring sharply-defined 2d rectangular regions of constant value.
Such a method is readily useful, for example, when making business decisions or executive actions that must be explained to a non-technical audience.  % JESSE: I think you could remove this sentence.
%CRISP is similar to classification and regression trees (CART), in that it partitions the feature space into contiguous blocks of constant value (``interpretable sharp partitions''), but was shown to lead to better performance.

Data-adaptive, interpretable sharp partitions are also useful in the creation of areal data from a set of spatial point-referenced data---turning a continuous spatial problem into a discrete one.  A common application of the framework arises when dividing a city, state, or other region into a set of contiguous cells, where values in each cell are aggregated to help anonymize individual demographic data.  Ensuring that the number and size of grid cells remains tractable, handling low-data regions, and preserving spatial structure are all important considerations for this problem. Ideally, one cell should contain data points which all map to a similar underlying value, and cell boundaries should represent significant change points in the value of the signal being estimated. If a cell is empty or contains a small number of data points, the statistical strength of its neighbors should be leveraged to both improve the accuracy of the reported areal data and further aid in anonymizing the cell which may otherwise be particularly vulnerable to deanonymization.  Viewed through this lens, we can interpret the areal-data creation task as a machine learning problem, one focused on finding sharp partitions that still achieve acceptable predictive loss.\footnote{We note that such a task will likely only represent a single step in a larger anonymization pipeline that may include other techniques such as additive noise and spatial blurring. While we provide no proofs of how strong the anonymization is for our method, we believe it is compatible with other methods that focus on adherence to a specified \textit{k}-anonymity threshold (e.g., \cite{cassa:etal:2006}).}

To this end, and motivated by the success of CRISP, we present MVTV, a method for interpretable, low-dimensional convex regression with sharp partitions. MVTV involves two main steps: (1) a novel maximum-variance heuristic to create a data-adaptive grid over the feature space; and (2) smoothing over this grid using a fast total variation denoising algorithm \cite{barbero:sra:2014}. The resulting model displays a good balance between interpretability, average accuracy, and degrees of freedom. We conduct a human study on the predictive interpretability of each method, showing both qualitatively and quantitatively that MVTV achieves superior performance over CART and CRISP.

% The remainder of this paper is organized as follows. Section \ref{sec:background} presents technical background on both CRISP and graph-based total variation denoising. In Section \ref{sec:algorithm}, we detail our algorithm and derive the gap statistic for both regression and classification scenarios. We then present a suite of benchmark experiments and human evaluations in Section \ref{sec:experiments} and conclude in Section \ref{sec:conclusion}.
% JESSE: I think you can safely remove this outline paragraph. The paper structure is pretty straightforward and doesn't deviate from expectations.

%% file: background.tex
% !TEX root = gaptv_icml.tex
\section{Background}
\label{sec:background}

% Both CRISP and GapTV formulate the interpretable regression problem as a regularized convex optimization problem. We first give a brief overview of the CRISP loss function and its computational complexity. We then give a brief preliminary overview of total variation denoising, the approach used by GapTV.

\subsection{Convex Regression with Interpretable Sharp Partitions}
\label{subsec:crisp}

\citet{petersen:etal:2016} propose the CRISP algorithm. As in our approach, they focus on the 2d scenario and divide the $(x_1, x_2)$ space into a grid via a data-adaptive procedure. For each dimension, they divide the space into $q$ regions, where each region break is chosen such that a region contains $1/q$ of the data. This creates a $q \times q$ grid of differently-sized cells, some of which may not contain any observations. A prediction matrix $M \in \R^{q \times q}$ is then learned, with each element $M_{ij}$ representing the prediction for all observations in the region specified by cell $(i,j)$.

CRISP applies a Euclidean penalty on the differences between adjacent rows and columns of $M$. The final estimator is then learned by solving the convex optimization problem,
\begin{equation}
\label{eqn:crisp_objective}
\begin{aligned}
& \underset{M \in \R^{q \times q}}{\text{minimize}}
& & 
\frac{1}{2}\sum_{i = 1}^n (y_i - \Omega(M, x_{1i}, x_{2i}))^2 + \lambda P(M) \, ,
\end{aligned}
\end{equation}
where $\Omega$ is a lookup function mapping $(x_{1i}, x_{2i})$ to the corresponding element in $M$. $P(M)$ is the group-fused lasso penalty on the rows and columns of $M$,
\begin{equation}
\label{eqn:crisp_penalty}
P(M) = \sum_{i = 1}^{q-1} \left[ \vnorm{M_{i\cdot} - M_{(i+1)\cdot}}_2 + \vnorm{M_{\cdot i} - M_{\cdot(i+1)}}_2 \right] \, ,
\end{equation}
where $M_{i\cdot}$ and $M_{\cdot i}$ are the $i^{\text{th}}$ row and column of $M$, respectively.

By rewriting $\Omega(\cdot)$ as a sparse binary selector matrix and introducting slack variables for each row and column in the $P(M)$ term, CRISP solves \eqref{eqn:crisp_objective} via ADMM.  % JESSE: I don't know what ADMM is. Do you need a citation here?
The resulting algorithm requires an initial step of $\mathcal{O}(n+q^4)$ operations for $n$ samples on a $q\times q$ grid, and has a per-iteration complexity of $\mathcal{O}(q^3)$. The authors recommend using $q=n$ when the size of the data is sufficiently small so as to be computationally tractable, and setting $q=100$ otherwise.

In comparison to other interpretable methods, such as CART and thin-plate splines (TPS), CRISP is shown to yield a good tradeoff between accuracy and interpretability.

\subsection{Graph-based Total Variation Denoising}
\label{subsec:graphtv}

Total variation (TV) denoising solves a convex regularized optimization problem defined generally over a graph $\mathcal{G} = (\mathcal{V}, \mathcal{E})$ with node set $\mathcal{V}$ and edge set $\mathcal{E}$,
\begin{equation}
\label{eqn:gfl_objective}
\begin{aligned}
& \underset{\boldsymbol\beta \in \R^{|\mathcal{V}|}}{\text{minimize}}
& & 
\sum_{s \in \mathcal{V}} \ell(\beta_s) + \lambda \sum_{(r,s) \in \mathcal{E}} |\beta_r - \beta_s| \, ,
\end{aligned}
\end{equation}
where $\ell$ is some smooth convex loss function over the value at a given node $\beta_s$. The solution to \eqref{eqn:gfl_objective} yields connected subgraphs (i.e. plateaus in the 2d case) of constant value. TV denoising has been shown to have attractive minimax rates theoretically \cite{wang:etal:2014} and is robust against model mispecification empirically, particularly in terms of worst-cell error \cite{tansey:scott:2016:multiscale}.

Many efficient, specialized algorithms have been developed for the case when $\ell$ is a Gaussian loss and the graph has a specific constrained form. For example, when $\mathcal{G}$ is a one-dimensional chain graph, \eqref{eqn:gfl_objective} is the ordinary (1d) fused lasso \cite{tibs:fusedlasso:2005}, solvable in linear time via dynamic programming \cite{johnson:2013}. When $\mathcal{G}$ is a d-dimensional grid graph, \eqref{eqn:gfl_objective} is typically referred to as total variation denoising \cite{rudin:osher:faterni:1992} or the graph-fused lasso, for which several efficient solutions have been proposed \cite{chambolle:darbon:2009,barbero:sra:2011,barbero:sra:2014}.

The TV denoising penalty was investigated as an alternative to CRISP in \cite{petersen:etal:2016}. They note anecdotally that TV denoising over-smooths when the same $q$ was used for both CRISP and TV denoising. We present a principled approach to choosing $q$ in a data-adaptive way that prevents over-smoothing and leads to a superior fit in terms of the accuracy-complexity tradeoff.

%% file: algorithm.tex
% !TEX root = gaptv_icml.tex
\section{The MVTV Algorithm}
\label{sec:algorithm}

We note that we can rewrite \eqref{eqn:crisp_objective} as a weighted least-squares problem,
\begin{equation}
\label{eqn:weighted_least_squares_objective}
\begin{aligned}
& \underset{\boldsymbol\beta \in \R^{q^2}}{\text{minimize}}
& & 
\frac{1}{2}\sum_{i = 1}^{q^2} \eta_i (\tilde{y}_i - \beta_i)^2 + \lambda g(\boldsymbol\beta) \, ,
\end{aligned}
\end{equation}
where $\boldsymbol\beta = \text{vec}(M)$ is the vectorized form of $M$, $\eta_i$ is the number of observations in the $i^{\text{th}}$ cell, and $\tilde{y}_i$ is the empirical average of the observations in the $i^{\text{th}}$ cell. $g(\cdot)$ is a penalty term that operates over a vector $\boldsymbol\beta$ rather than a matrix $M$.

We choose $g(\cdot)$ to be a graph-based total variation penalty,
\begin{equation}
\label{eqn:graph_tv_penalty}
g(\boldsymbol\beta) = \sum_{(r,s) \in \mathcal{E}} |\beta_r - \beta_s| \, ,
\end{equation}
where $\mathcal{E}$ is the set of edges defining adjacent cells on the $q \times q$ grid graph. Having formulated the problem as a graph TV denoising problem, we can now use the convex minimization algorithm of \citet{barbero:sra:2014} (or any other suitable algorithm) to efficiently solve \eqref{eqn:weighted_least_squares_objective}.

We auto-tune the two hyperparameters: $q$, the granularity of the grid, and $\lambda$, the regularization parameter. We take a pipelined approach by first choosing $q$ and then selecting $\lambda$ under the chosen $q$ value.

\subsection{Choosing bins via a maximum variance heuristic}
\label{subsec:choosing_q}
The recommendation for CRISP is to choose $q = n$, assuming the computation required is feasible. Doing so creates a very sparse grid, with $q-1 \times q$ empty cells. However, by tying together the rows and columns of the grid, each CRISP cell actually draws statistical strength from a large number of bins. This compensates for the data sparsity problem and results in reasonably good fits despite the sparse grid.

Choosing $q = n$ does not work for our TV denoising approach. Since the graph-based TV penalty only ties together adjacent cells, long patches of sparsity overwhelm the model and result in over-smoothing. If one instead chooses a smaller value of $q$, however, the TV penalty performs quite well. The challenge is therefore to adaptively choose $q$ to fit the appropriate level of overall data sparsity. We do this by choosing the grid which maximizes the sum of variances of all cells:
\begin{equation}
\label{eqn:max_variance}
q = \underset{q}{\text{argmax}}\quad \sum_{c \in \mathcal{C}(q)} \hat{\text{var}}(\mathbf{y}_c) \, ,
\end{equation}
where $\mathcal{C}(q)$ is the set of cells in the $q \times q$ grid and $\text{var}(\emptyset) = 0$.

Choosing the grid is a tradeoff between each cell's fit to the data and the total number of cells. Each sample $y_i$ is assumed to be IID conditioned on being in the same cell. We find that maximizing the sum of variances as in \eqref{eqn:max_variance} serves as a useful heuristic for finding cells that fit well to the distribution of the data and prevent overfitting by using too many cells. 

A clear connection also exists between our heuristic and principal components analysis (PCA). Since we are dealing with univariate observations, maximizing the variance corresponds to finding the approximation to the first principal component of the data. The TV penalty then helps to smooth over these principal components by incorporating the spatial adjacency information. Such a connection presents the possiblity for future extensions to multivariate observations and smoothing using group TV methods like the network lasso \citep{hallac:etal:2015}.

Once a value of $q$ has been chosen, $\lambda$ can be chosen by following a solution path approach. For the regression scenario with a Gaussian loss, as in \eqref{eqn:weighted_least_squares_objective}, determining the degrees of freedom is well studied \citep{tibs:taylor:2011}. Thus, we could select $\lambda$ via an information criterion such as AIC or BIC. We select $\lambda$ via cross-validation because we found empirically that it produces better results.

%% file: evaluation.tex
% !TEX root = gaptv_icml.tex
\section{Case Study: Austin Crime Data}
\label{sec:eval}
\begin{figure}[t]
\centering
\begin{subfigure}[t]{0.23\textwidth}
\includegraphics[width=\textwidth]{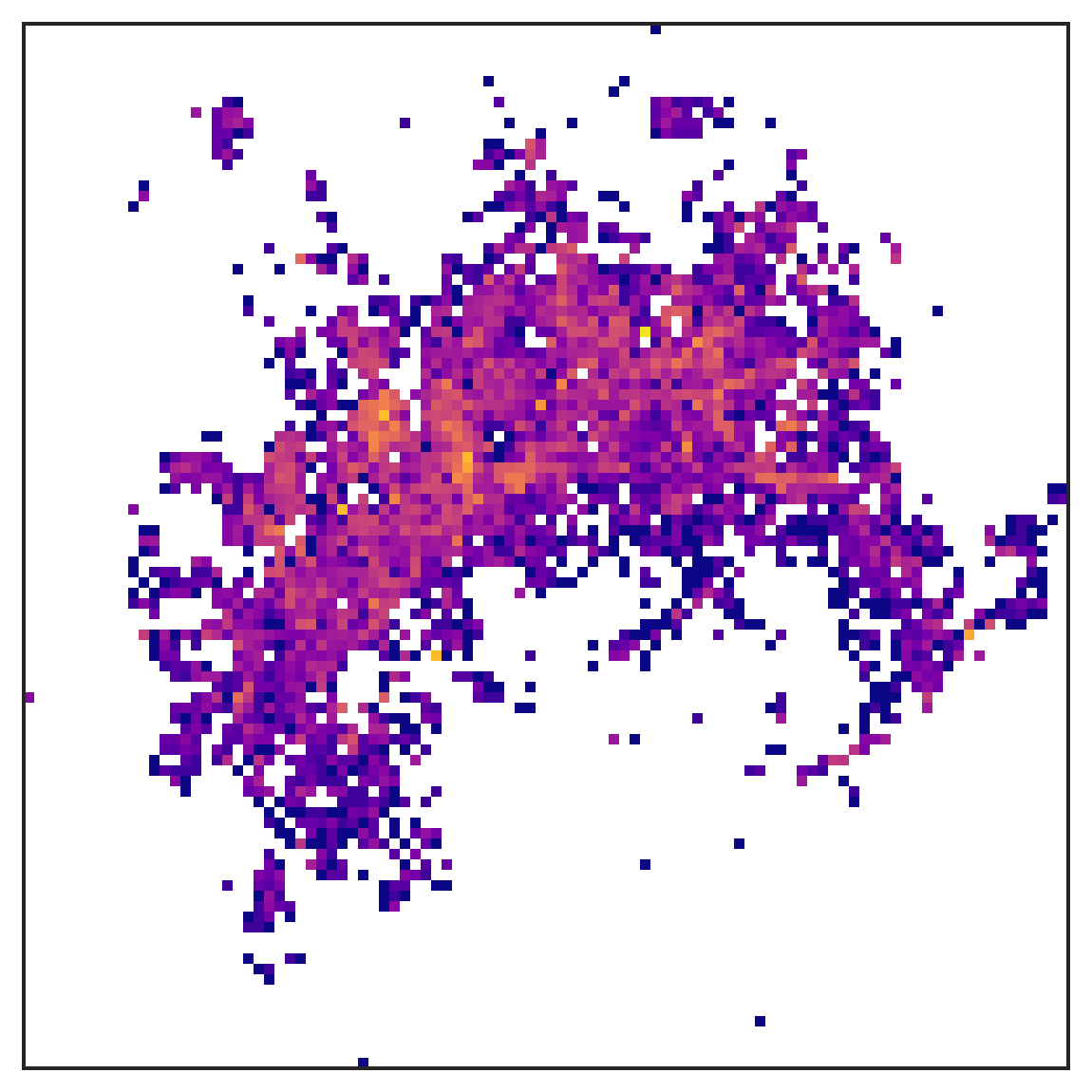}
\caption{Raw}
\end{subfigure}
\begin{subfigure}[t]{0.23\textwidth}
\includegraphics[width=\textwidth]{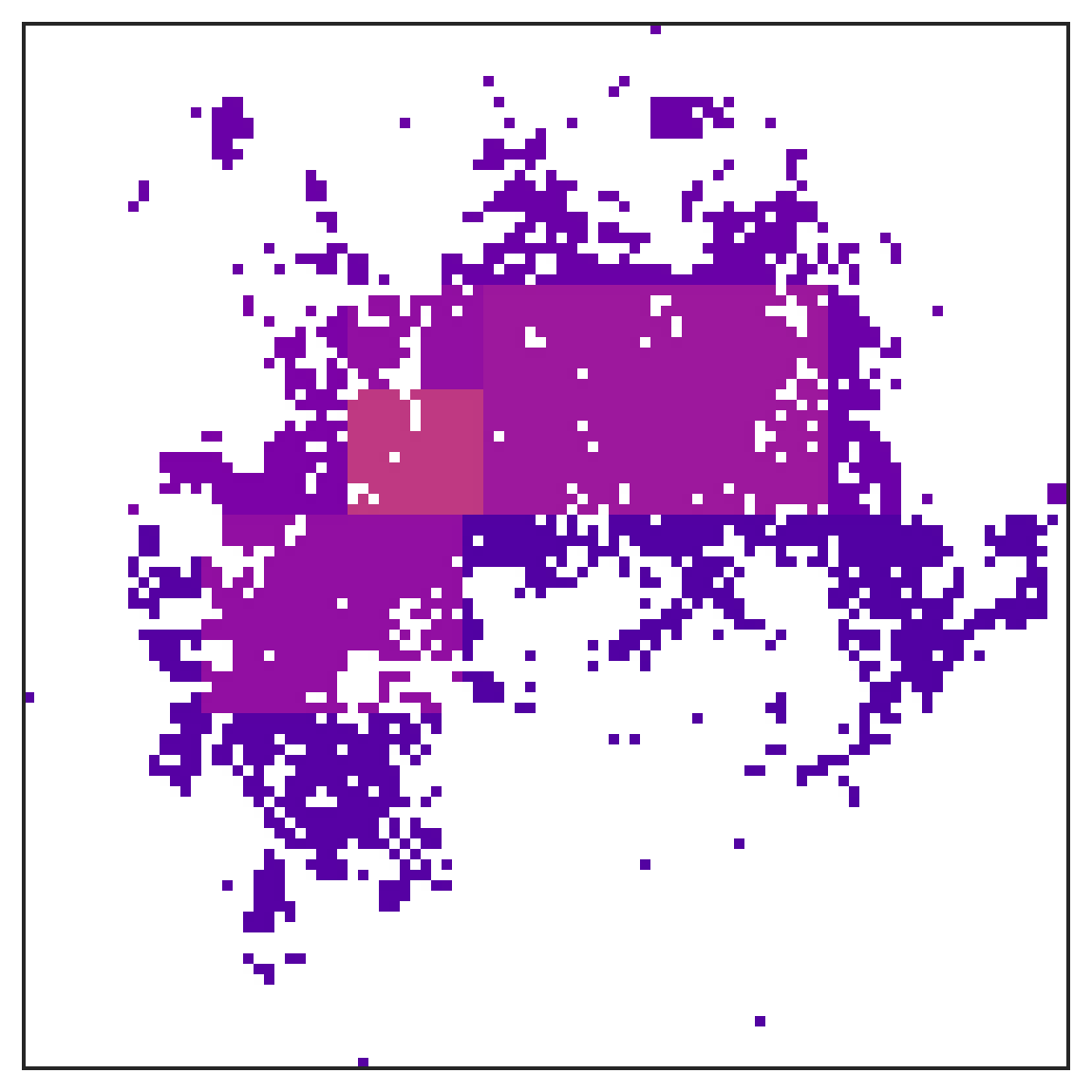}
\caption{CART}
\end{subfigure}
\begin{subfigure}[t]{0.23\textwidth}
\includegraphics[width=\textwidth]{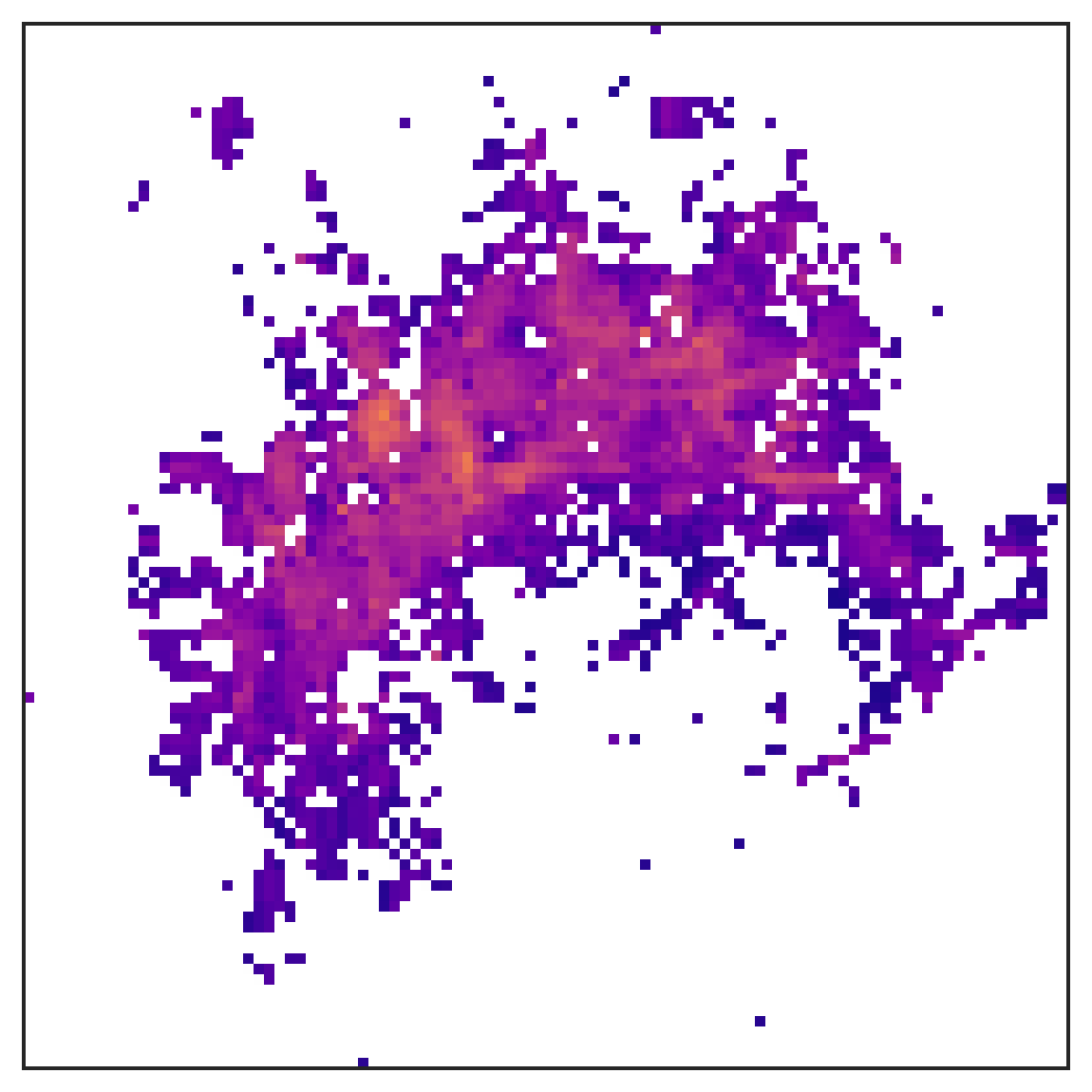}
\caption{CRISP}
\end{subfigure}
\begin{subfigure}[t]{0.23\textwidth}
\includegraphics[width=\textwidth]{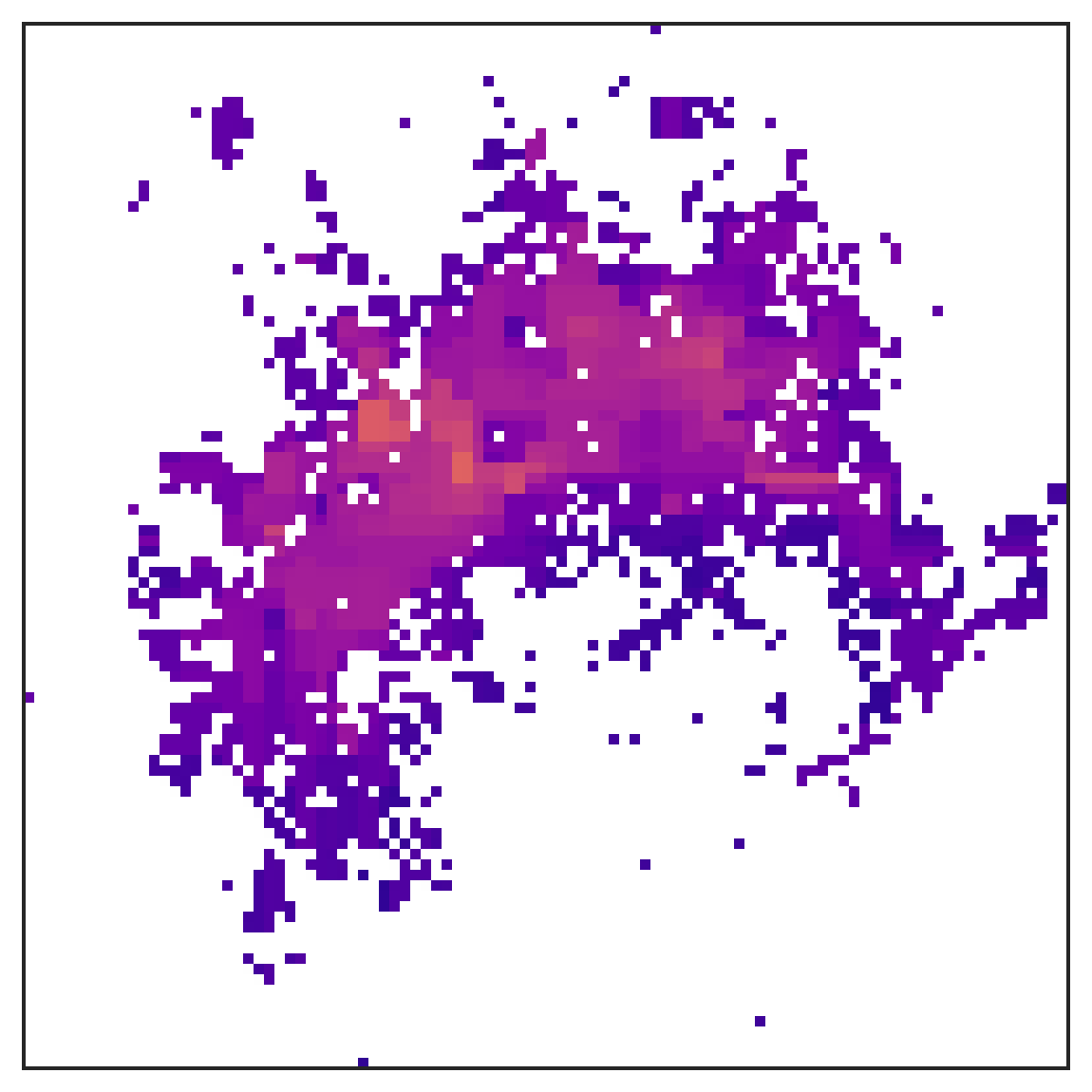} 
\caption{MVTV}
\end{subfigure}
\caption{\label{fig:crime_results} Areal data results for the Austin crime data. The maps show the raw fine-grained results (Panel A) and the results of the three main methods. Qualitatively, CART (Panel B) over-smooths and creates too few regions in the city; CRISP (Panel C) under-smooths, creating too many regions; and MVTV (Panel D) provides a good balance that yields interpretable sections.
\vspace{-0.1in}}
\end{figure}

We applied CART, CRISP, and MVTV to a dataset of publicly-available crime report counts\footnote{\url{https://www.data.gov/open-gov/}} in Austin, Texas in 2014. To preprocess the data, we binned all observations into a fine-grained $100 \times 100$ grid based on latitude and longitude, then took the log of the total counts in each cell. Points with zero observed crimes were omitted from the dataset as it is unclear whether they represented the absence of crime or a location outside the boundary of the local police department. Figure \ref{fig:crime_results} (Panel A) shows the raw data for Austin.

The MVTV method used $q$ values in the range $[2,100]$ and the CRISP method used $q=100$. We ran a 20-fold cross-validation to measure RMSE and calculated plateaus with a fully-connected grid (i.e., as if all pixels were connected) which we then projected back to the real data for every non-missing point. Figure \ref{fig:crime_results} shows the qualitative results for CART (Panel B), CRISP (Panel C), and MVTV (Panel D). The CART model clearly over-smooths by dividing the entire city into huge blocks of constant plateaus; conversely, CRISP under-smooths and creates too many regions. The MVTV method finds an appealing visual balance, creating flexible plateaus that partition the city well. These results are confirmed quantitatively in Table \ref{tab:crime_results}, where MVTV outperforms the other methods in terms of AIC.

To evaluate the interpretability of the MVTV method against the benchmark CART and CRISP methods, we ran a Mechanical Turk study with human annotators.
The annotation task was to choose a grayscale value for a held-out cell in the center of a $7\times 7$ patch of data.
Each annotator was shown a patch as rendered by MVTV, CART, CRISP, and as raw data; each task involved two randomly sampled patches from the Austin crime dataset ($4\times 2=8$ patches per annotator, shown in random order).

\begin{table}
\centering
\begin{tabular}{|l|cc|}
\hline
\multicolumn{1}{|c}{} & \multicolumn{2}{|c|}{Austin Crime Data} \\ \hline
 & AIC & Human error $\times 10^{-2}$ \\ \hline
CART  & 11139.29 & 3.24$\pm$0.341 \\
CRISP & 18326.33 & 3.99$\pm$0.664 \\
MVTV & {\bf 10327.58} & {\bf 2.75}$\pm$0.334 \\
\hline
\end{tabular}
\caption{\label{tab:crime_results} Results for the three methods on crime data for Austin.
The MVTV method achieves the best trade-off between accuracy and the number of constant regions, as measured by AIC. Human annotator predictions are also statistically significantly closer than when annotators are shown raw data, which neither CART nor CRISP achieve.
\vspace{-0.2in}}
\end{table}

We added two additional uniform color validation patches.
We gathered information from $207$ annotators for $190$ patches, discarding $37$ annotators who were not within 10\% of the uniform value in the solid-colored validation patches.
We measured the squared difference between the average annotators' predictions per (patch, method) combination against the true value in the raw data, shown in Table~\ref{tab:crime_results} (rightmost column).

The raw data is noisy and has high local variance, and so annotators do poorly at the prediction task without any smoothing ($0.0471\pm0.00539$, not shown in Table~\ref{tab:crime_results}).
The over-smoothed CART values create too many uniform plateaus where the annotators cannot reasonably predict anything other than the missing uniform value, which has low accuracy.
The CRISP method fails to sufficiently smooth the data, resulting in overly noisy patches which again makes the prediction task difficult.
MVTV provides a good balance of smoothing and flexibility.

According to a Tukey's range test comparing pairwise human annotations across methods, MVTV statistically significantly outperforms the raw data for the human prediction task; by contrast, CART and CRISP fail to outperform the raw data.
No smoothing methods were shown to outperform one another with significance.